\documentclass[10pt,twocolumn,letterpaper]{article}

\usepackage[pagenumbers]{cvpr} 
\usepackage{adjustbox}
\usepackage[T1]{fontenc}
\usepackage{tikz}
\usepackage{url}            
\usepackage{booktabs}       
\usepackage{amsfonts}       
\usepackage{nicefrac}       
\usepackage{microtype}      
\usepackage{xcolor}         
\usepackage{array}

\usepackage{tikz}
\usetikzlibrary{tikzmark,calc}
\usepackage{xcolor}

\definecolor{deltapink}{RGB}{235,95,145}

\ExplSyntaxOn
\seq_new:N \g_pp_annotations_seq
\int_new:N \g_pp_mark_int

\NewDocumentCommand{\ppmark}{mm}
{
  \int_gincr:N \g_pp_mark_int

  \tikzmarknode{pp-\int_use:N \g_pp_mark_int}{#1}

  \seq_gput_right:Nx \g_pp_annotations_seq
  {
    \exp_not:N \node [
      anchor=base~west,
      inner~sep=0pt,
      outer~sep=0pt,
      text=deltapink,
      font=\exp_not:N \fontsize{6.3pt}{6pt}
           \exp_not:N \selectfont
    ]
    at ([xshift=0.5pt,yshift=-1.3pt]
        pp-\int_use:N \g_pp_mark_int .base~east)
    {\exp_not:n{#2}};
  }
}

\NewDocumentCommand{\ppflush}{}
{
  \tikz[remember~picture,overlay]{
    \seq_use:Nn \g_pp_annotations_seq {~}
  }
  \seq_gclear:N \g_pp_annotations_seq
}
\ExplSyntaxOff

\usepackage{epsfig}
\usepackage{graphicx}
\usepackage{amsmath}
\usepackage{tabularx}
\usepackage{amssymb}
\usepackage{amsthm}
\usepackage{paralist}
\usepackage{pifont}
\usepackage{multirow}
\usepackage{bm}
\usepackage{bbm}
\usepackage{diagbox}
\usepackage{makecell}
\usepackage{float}
\usepackage{stfloats}
\usepackage{algorithm}  
\usepackage{color, colortbl}
\usepackage{arydshln}
\usepackage{marvosym}
\usepackage{enumitem}
\usepackage{subcaption}
\usetikzlibrary{arrows.meta,calc,fit,positioning}
\usepackage{algpseudocode}

\theoremstyle{definition}

\theoremstyle{plain}

\renewcommand{\eqref}[1]{Eq.~(\ref{#1})}

\newcommand{\gr}{\rowcolor[gray]{.95}} 
\newcommand{\good}{\rowcolor[RGB]{234,243,233}}
\definecolor{cvprblue}{rgb}{0.21,0.49,0.74}
\usepackage[pagebackref,breaklinks,colorlinks,allcolors=cvprblue]{hyperref}

\def\confName{CVPR}
\def\confYear{2027}

\title{3DVLA: Enhancing Vision-Language-Action Models via 3D Spatial and Instance Understanding}
\author{
Zhongyu Xia$^{1,2}$\textsuperscript{*}\quad Yousen Tang$^1$\textsuperscript{†*}\quad Bingqing Wei$^1$\quad Yongtao Wang$^{1,2}$\textsuperscript{\Letter}
\\
$^1$ Wangxuan Institute of Computer Technology, Peking University
\quad 
$^2$ VGI Labs Co., Ltd.
\\
{\tt\small \{xiazhongyu,wyt\}@pku.edu.cn
\quad \tt\small tangyousen@tongji.edu.cn
\quad \tt\small bingqing.wei@stu.pku.edu.cn
}
}

\begin{document}

\maketitle

\renewcommand{\thefootnote}{}
\footnotetext{
\textsuperscript{*} Equal contribution. ~\textsuperscript{\Letter} Corresponding author.
\textsuperscript{†} This work was done as an intern at PKU.
}

\begin{abstract}

Vision-Language-Action models have achieved remarkable progress in robotic manipulation, yet they suffer from a critical limitation: a lack of 3D scene understanding. This deficiency manifests as three intertwined challenges: weak extraction of 3D spatial positions without enforcing multi-view consistency, inadequate 3D instance understanding, and fragile reasoning under occlusion. Although mature 3D perception methods exist, their direct integration into VLA pipelines is hindered by architectural incompatibility and by heavy reliance on costly instance-level annotations. To address these challenges, we propose 3DVLA, a framework compatible with existing VLA models, which introduces 3D understanding without requiring extra manual labels or discarding VLM priors. Specifically, 3DVLA tackles the three challenges through: (1) pervasive 3D feature encoding with explicit multi-view consistency constraints across all modalities and a Spatially-Conditioned Geometry Aggregation method, (2) an object-centric 3D instance module with high-level instance tokens for 3D instance awareness, and (3) a masked self-supervised 3D encoding branch that retains its predictor for visual token completion to handle occlusions. We integrate 3DVLA with multiple VLA baselines and evaluate on LIBERO-Plus, RoboTwin 2.0, and real-world settings. Results show consistent and significant gains in manipulation performance, validating both the effectiveness and plug-and-play compatibility of our approach.

\end{abstract}

\section{Introduction}
\label{sec:intro}

By leveraging large-scale vision-language pre-training, Vision-Language-Action Models (VLAs)~\cite{rt1,rt2,openvla,octo} have demonstrated impressive manipulation capabilities across diverse tasks and embodiments, underscoring their immense practical value for general-purpose robotics. 
However, most pretrained VLA policies represent visual observations primarily as image tokens, without explicitly maintaining geometric relationships across camera views or persistent 3D identities for physical objects.
This deficiency impedes their deployment in real-world environments where spatial reasoning is crucial and can be characterized by three tightly coupled challenges.

\textbf{The first challenge is weak 3D spatial position extraction and the absence of multi-view consistency constraints.}
Existing VLA architectures typically encode single-view observations or loosely fuse multiple views without explicitly reconstructing 3D coordinates or enforcing consistencies across viewpoints. 
As a result, the policy cannot ground its decisions in a metric 3D reference frame, making it incapable of reasoning about precise 6-DoF poses.

\textbf{The second challenge is insufficient 3D instance understanding.}
Standard VLAs reason about objects as flat image regions, without access to their volumetric extent, 3D boundaries, or fine-grained geometry. 
Consequently, they struggle with tasks that demand precise spatial interaction (such as grasping a specific part of an object or placing one item relative to another’s 3D structure) because the model cannot reliably distinguish which pixels belong to which object instance in 3D space. 

Recent work~\cite{gam,3dthinkvla,imvoxelnet,r4det, openad, henet++} has explored different ways of incorporating 3D information into VLA policies while leveraging pretrained vision-language representations. However, providing geometric inputs alone does not necessarily establish consistent associations between physical instances across camera views, nor does it provide useful visual features for regions that are not directly observed. Integrating specialized 3D encoders forces a redesign that discards this pre-training, sacrificing the generalization and language grounding that make VLAs valuable. We therefore focus on organizing multi-view geometric evidence into object-centric representations and providing action-relevant features at candidate locations under partial observation without requiring extra manual labels or discarding VLM priors.


\textbf{The third challenge is fragile reasoning under occlusion, particularly when viewpoint variations hide task-relevant object parts.}
A telling example is observed on the LIBERO-Plus benchmark~\cite{liberoplus}: slightly shifting the camera viewpoint so that the target object becomes partially occluded causes the state-of-the-art VLAs' success rates to drop dramatically. This brittleness arises because purely 2D models lack the capacity to infer the unseen geometry and maintain a coherent spatial memory, leading to action generation based on incomplete and misleading visual cues. 
This fragility calls for a mechanism that can explicitly infer unseen geometry from partial views. Interestingly, masked self-supervised learning offers an overlooked building block.
Masked self-supervised methods~\cite{ijepa} typically discard the predictor used to predict masked tokens after self‑supervised training. 
However, we realize that after adopting a unified 3D positional encoding, this predictor can be repurposed to complete tokens at occluded locations.

\begin{figure*}[t]
\centering
\includegraphics[width=\textwidth]{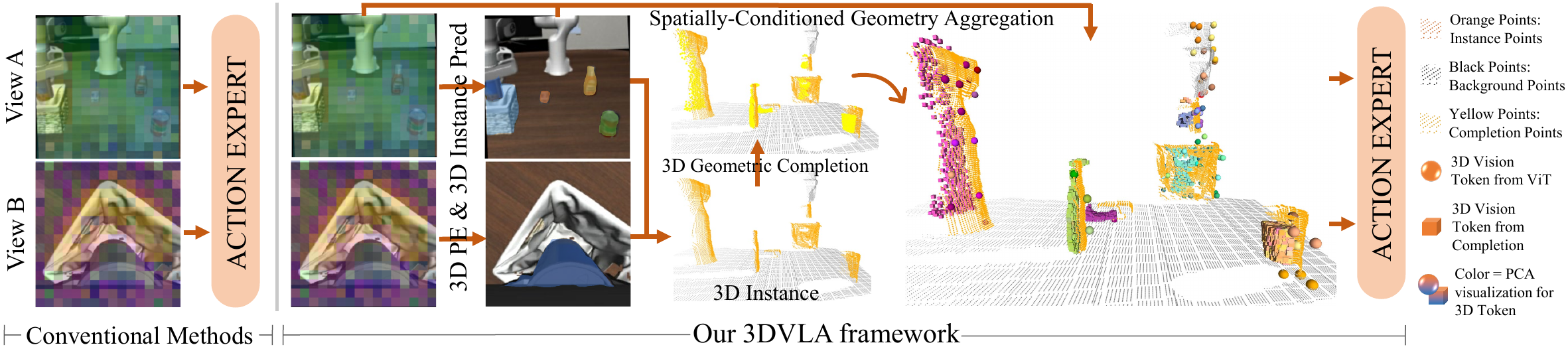}
\caption{\textbf{Overview of 3DVLA.} Compared to conventional methods, 3DVLA extracts view-consistent 3D instances (orange) and synthesizes occluded geometries (yellow). These features are aggregated into spatially-grounded vision tokens to enable robust manipulation under complex occlusions.}
\label{fig:intro}
\end{figure*}

In this paper, we propose 3DVLA, a plug-and-play method that enriches VLA models with robust 3D understanding by leveraging priors from off-the-shelf foundation perception models. 
Crucially, 3DVLA is designed to be fully compatible with VLM pre-training and requires no extra manual annotations. 
Concretely, our contributions include:

\textbf{(1)} To address the first challenge, we impose~\textit{Multi-View Spatial Fusion} and~\textit{Spatially-Conditioned Geometry Aggregation}, enforcing explicit multi-view consistency constraints throughout the entire model. 

\textbf{(2)} To address the second challenge of weak 3D instance understanding, 3DVLA incorporates an~\textit{Object-Centric 3D Instance Module} and introduces high-level 3D instance tokens that explicitly encode object identity and geometry. 

\textbf{(3)} For the third challenge of occlusion reasoning, we complement the architecture with a 3D encoding branch trained via masked self-supervision. Unlike conventional masked auto-encoding paradigms that discard the lightweight predictor after pre-training, we retain it and repurpose it for visual token completion, enabling the model to reconstruct occluded visual features and reason about invisible parts of objects during policy execution. 

\textbf{(4)} We integrate 3DVLA with multiple VLA baselines and conduct experiments on the LIBERO-Plus and RoboTwin 2.0 benchmarks. 
The results demonstrate that our method consistently improves manipulation performance across different architectures and tasks, validating both its effectiveness and its plug-and-play compatibility.

\section{Related Work}
\label{sec:related}




\noindent\textbf{Vision-Language-Action Models} 
With the development of large language models and large-scale robotic datasets, Vision-Language-Action (VLA) models have emerged as a crucial paradigm for general-purpose robot policy learning. 
OpenVLA~\cite{openvla} fuses DINOv2 and SigLIP, while NORA~\cite{nora} is built upon a lightweight VLM.
WorldVLA~\cite{worldvla} unifies action and future-state modeling into an autoregressive world model, and UniVLA~\cite{univla} enables cross-embodiment generalization via a task-centric implicit action space. 
$\pi_0$~\cite{pi_0} pioneered flow matching for continuous action generation, while $\pi_{0.5}$~\cite{pi_05} added open-world generalization through heterogeneous co-training and knowledge insulation. 
RIPT-VLA~\cite{riptvla} extends the pre-training plus supervised fine-tuning pipeline with a reinforcement-learning-based interactive post-training stage. 
Diffusion Policy (DP)~\cite{diffusion} formulates action generation as a denoising diffusion probabilistic process, while DP3~\cite{dp3} replaces its 2D input with 3D point cloud representations. 
ACT~\cite{act} recasts action prediction as a conditional VAE-driven sequence-generation problem, and RDT~\cite{rdt} employs a physically interpretable, unified action space within a scalable diffusion Transformer.

Recent work has investigated various approaches to incorporating geometric information into robot policies. 3DThinkVLA~\cite{3dthinkvla} transfers 3D geometric and spatial reasoning priors into a VLA through latent distillation and co-training. SG-WAM~\cite{sgwam} introduces geometry-aware supervision and action-conditioned latent future prediction in the policy representation space. The Geometric Action Model (GAM)~\cite{gam} repurposes a pretrained geometric foundation model for joint future geometry and action prediction.
These approaches highlight the benefits of geometric priors, structured spatial representations, and predictive modeling for manipulation. Our work focuses on a complementary problem: establishing object-level associations across multiple views in a shared 3D coordinate frame and augmenting these representations with latent visual features at candidate unobserved 3D locations. To this end, 3DVLA combines cross-view instance modeling with coordinate-conditioned feature completion and spatially guided aggregation for downstream action prediction.


\noindent\textbf{Self-Supervised Representation Learning}
In the construction of visual encoders, self-supervised pre-training is a commonly adopted technique.
Early methods~\cite{ilpo, lapa} learn latent actions by modeling pixel variations between adjacent frames, but they are prone to degenerating into reliance on local visual correlations.
DINOv2~\cite{dinov2} employs masked self-distillation for Vision Transformer pre-training.
I-JEPA~\cite{ijepa} transitioned from pixel-level reconstruction to latent space modeling. 
This paradigm has further influenced the field of embodied learning; for example, VLA-JEPA \cite{vlajepa} extends it to action-relevant representation alignment tasks. 
However, these self-supervised methods typically discard the masked tokens predictor after training. 
We realize that after adopting a unified 3D positional encoding, this predictor can be repurposed to complete tokens at occluded locations.

\begin{figure*}[t]
\centering
\includegraphics[width=0.9\textwidth]{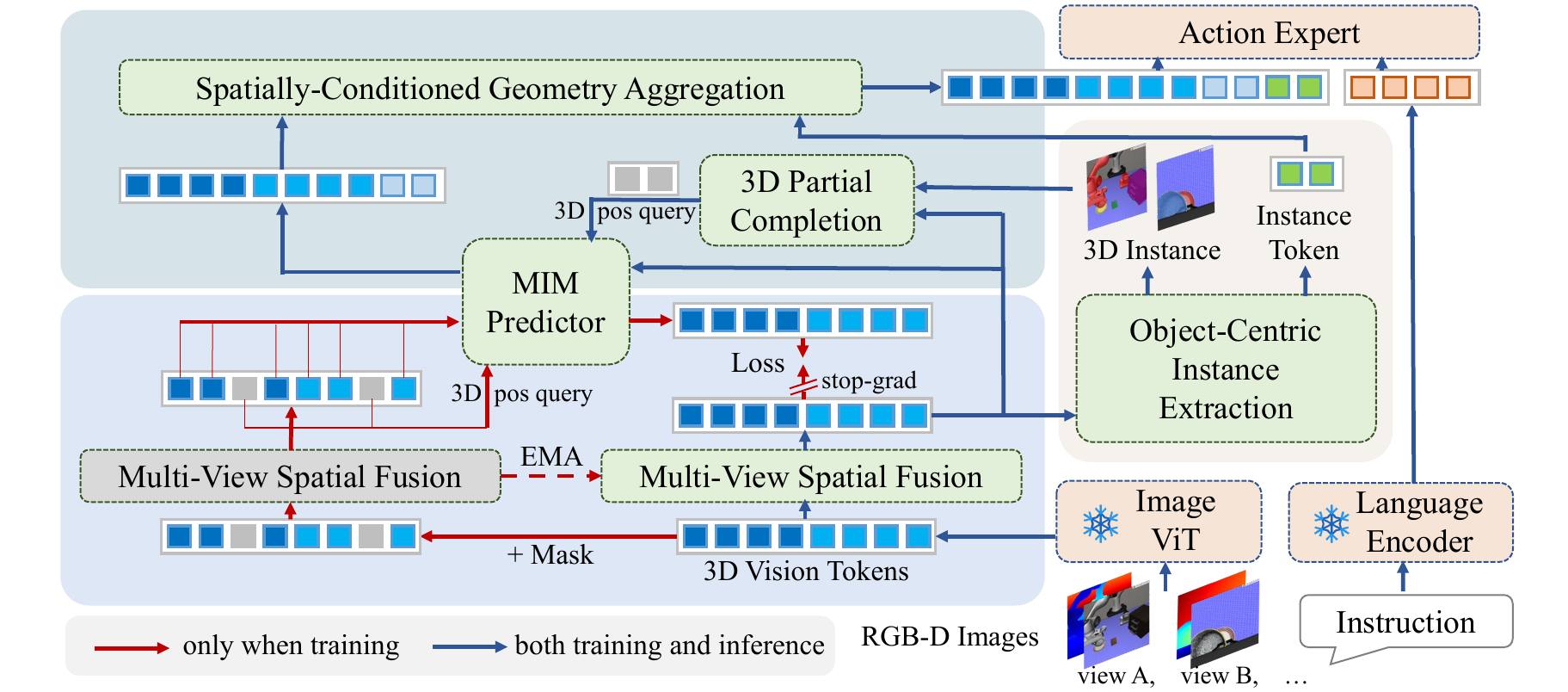}
\caption{\textbf{Overall Architecture of 3DVLA.} The pipeline consists of four interconnected components: (1) Multi-View Spatial Fusion, (2) Object-Centric 3D Instance Module, (3) Coordinate-Driven 3D Self-Supervised Predictor, and (4) Spatially-Conditioned Geometry Aggregation.}
\label{fig:overall}
\end{figure*}

\section{Method}
\label{sec:method}

We introduce 3DVLA, a 3D framework that enhances 2D Vision-Language-Action (VLA) models without disrupting their pre-trained representations. As illustrated in Figure~\ref{fig:overall}, 3DVLA consists of four interconnected components: (1) \textit{Multi-View Spatial Fusion} lifts 2D visual tokens into a unified 3D memory bank to enforce multi-view consistency; (2) The \textit{Object-Centric 3D Instance Module} extracts physical entities directly in 3D space; (3) A \textit{Coordinate-Driven 3D Self-Supervised Predictor} synthesizes occluded geometric structures; (4) \textit{Spatially-Conditioned Geometry Aggregation} explicitly embeds these 3D geometries as affordance-aware instance tokens for the downstream action expert.

\subsection{Multi-View Spatial Fusion}
\label{subsec:spatial_fusion}

To construct a unified 3D representation from multi-view feature maps $\mathcal{X}_{fmap} \in \mathbb{R}^{V \times C \times H \times W}$ without computationally expensive voxelization, we flatten the spatial dimensions into an unordered sequence $\mathcal{M}_{flat} \in \mathbb{R}^{N \times C}$ ($N = V \times H \times W$). Using the unprojected 3D coordinate grids $\mathbf{P}_{flat} \in \mathbb{R}^{N \times 3}$, we ground the visual features in absolute 3D space via a coordinate projection:
\begin{equation}
\begin{split}
\mathbf{E}_{3D} = \operatorname{LN}\Big( \mathbf{W}_2 \operatorname{GELU}(\mathbf{W}_1 \mathbf{P}_{flat} + \mathbf{b}_1) + \mathbf{b}_2 \Big),
\end{split}
\end{equation}

where $\operatorname{LN}(\cdot)$ denotes Layer Normalization, $\mathbf{W}_{(\cdot)}$ and $\mathbf{b}_{(\cdot)}$ denote learnable weights and biases. The position-aware features ($\mathcal{M}_{flat} + \mathbf{E}_{3D}$) are processed by a multi-layer Transformer Encoder. Driven by absolute spatial embeddings, this module natively establishes cross-view geometric correspondences, yielding a globally consistent memory $\mathcal{M}$.
While the fusion module relies on absolute coordinates to align global context, synthesizing occluded structures (Section \ref{subsec:ssl_predictor}) strictly requires fine-grained relative distances. To provide this, we apply Continuous 3D Rotary Position Embedding (RoPE) exclusively to the attention queries and keys. For a specific axis component $p \in \{x, y, z\}$ of a 3D coordinate and its channel dimension $d$, the phase angle is computed continuously as $\theta_{d} = p \cdot 10000^{-2d/D_{axis}}$, where $D_{axis}$ is the channel dimension allocated per spatial axis. The independent embeddings for the three axes are concatenated to impose strict Euclidean relative-distance priors on the network.

\subsection{Object-Centric 3D Instance Module}
\label{subsec:instance_module}

To overcome the limitations of 2D bounding boxes and masks, we initialize state probes directly in continuous 3D space. This unifies absolute 3D geometric anchoring with semantic feature extraction. 

\subsubsection{3D Probe Decoding and Multi-Scale Sampling}
\label{subsec:sampling}

To extract entities from the globally consistent memory $\mathcal{M}$ (Section \ref{subsec:spatial_fusion}), we instantiate $N_q$ state probes $\mathcal{Q} = \{ (\mathbf{c}_j, \mathbf{p}_j) \}_{j=1}^{N_q}$, where $\mathbf{c}_j \in \mathbb{R}^C$ encapsulates the semantic state and $\mathbf{p}_j \in \mathbb{R}^3$ denotes its 3D reference point. Importantly, each probe represents an instance through both its 3D reference point and its learned semantic state. The semantic feature is refined by aggregating multi-scale observations across valid camera views, while view-conditioned bounding-box and mask predictions provide complementary cues about the instance's appearance and spatial extent. Specifically, $\mathcal{M}$ is decoded via transposed convolutions into a multi-scale semantic pyramid $\mathcal{M}_{ms}$ (strides 8, 16, 32) and a 1/4-scale map $\mathcal{M}_{mask}^{(4)}$ for high-resolution mask supervision.

The 3D point $\mathbf{p}_j$ is projected to the $v$-th camera view via intrinsic $\mathbf{K}_v$ and extrinsic $[\mathbf{R}_v | \mathbf{t}_v]$ matrices. To prevent numerical instability near the camera plane, we introduce a depth clamping mechanism:
\begin{equation}
\mathbf{u}_{j,v} = \Phi_{norm}\left( \frac{\mathbf{K}_v (\mathbf{R}_v \mathbf{p}_j + \mathbf{t}_v)}{\max(Z_c, \epsilon)} \right) \in [0, 1]^2,
\end{equation}
where $Z_c$ is the projected depth, $\epsilon = 0.01$ serves as a numerical stabilizer, and $\Phi_{norm}$ normalizes the coordinates. Treating $\mathbf{u}_{j,v}$ as the sampling pivot, the semantic vector $\mathbf{c}_j$ predicts 2D spatial offsets $\Delta \mathbf{u}$ and attention weights $A$. The probe then extracts textures from $\mathcal{M}_{ms}$ via Deformable Cross-Attention:

\begin{equation}
\tilde{\mathbf{u}}_{j,v,m,k,s} = \mathbf{W}' \mathcal{M}_{ms}^{(s)}\left(\mathbf{u}_{j,v} + \Delta \mathbf{u}_{j,v,m,k,s}\right),
\label{eq:offset_feature}
\end{equation}
\begin{equation}
\mathbf{h}_{j,v} = \sum_{m=1}^M \mathbf{W}_m \sum_{k=1}^K \sum_{s=1}^S A_{j,v,m,k,s} \cdot \tilde{\mathbf{u}}_{j,v,m,k,s},
\label{eq:main_aggregation}
\end{equation}
where $M, K, S$ denote the number of attention heads, sampled keys, and feature scales, respectively, and the predicted attention weights $A$ are normalized to sum to 1.

Crucially, a single 3D physical entity naturally projects to drastically different 2D shapes across multiple cameras. Forcing a shared 3D semantic state $\mathbf{c}_j$ to directly predict these varying 2D masks would lead to conflicting gradient updates. To resolve this, we strictly anchor predictions to $\mathbf{p}_j$, but condition the regression of view-dependent bounding boxes and masks on a view-specific positional encoding $\operatorname{PE}(\mathbf{u}_{j,v})$. This enables the shared probe $\mathbf{c}_j$ to handle perspective variations without compromising its 3D topological consistency.

\subsubsection{Occlusion-Aware Gating and Bounded Coordinate Refinement}
\label{subsubsec:refinement}

To filter geometric noise from occluded or out-of-view perspectives, an adaptive gating mechanism evaluates view validity: $g_{j,v} = \operatorname{sigmoid}( \operatorname{MLP}(\mathbf{h}_{j,v}) /\tau_{view} )$, where $\tau_{view}$ is a learnable temperature scalar that controls the sharpness of the gating. At each decoder layer $l$, the semantic state is residually updated via $\mathbf{c}_j^{(l+1)} = \mathbf{c}_j^{(l)} + \sum_v \tilde{g}_{j,v} \mathbf{h}_{j,v}^{(l)}$, where $\tilde{g}_{j,v}$ is the normalized gating weight, naturally suppressing unclear observations.

Concurrently, the reference point $\mathbf{p}_j$ must be calibrated to correct initial depth inaccuracies. To prevent spatial divergence during training, we propose a bounded coordinate refinement:
\begin{equation}
\mathbf{p}_j^{(l+1)} = \mathbf{p}_j^{(l)} + \alpha \cdot \tanh\Big( \operatorname{MLP}_{coord}(\mathbf{c}_j^{(l+1)}) \Big).
\end{equation}
The $\tanh$ activation, scaled by a predefined physical factor $\alpha$, limits the maximum spatial displacement per layer, guaranteeing stable 3D centroid convergence.

\subsubsection{Global 3D Matching and Joint Optimization}
\label{subsubsec:matching_loss}

We shift label assignment exclusively into the 3D space to eliminate cross-view ID conflicts. Pseudo ground-truth 2D bounding boxes and masks are generated by an off-the-shelf frozen perception model~\cite{vlsam2}, and their 3D centroids are derived by unprojecting 2D centers using aligned depth maps. The global matching cost $\mathcal{C}_{global}(j, k)$ between probe $j$ and target $k$ is averaged across all valid views $\mathcal{V}_k$:
\begin{equation}
\mathcal{C}_{global}(j, k) = \frac{1}{\max(|\mathcal{V}_k|, 1)} \sum_{v \in \mathcal{V}_k} \mathcal{C}_v(j, k),
\end{equation}
where $\mathcal{C}_v(j,k)$ measures the semantic and geometric consistency between prediction $j$ and target $k$ in view $v$, including classification, bounding-box, mask, and 3D centroid distance terms. We apply the Hungarian algorithm once to $\mathcal{C}_{global}$ to obtain the optimal assignment.

The network is optimized by 6 joint loss components:
\begin{equation}
\label{eq:joint_loss}
\begin{split}
\mathcal{L} &= \lambda_{cls}\mathcal{L}_{cls} + \lambda_{box}\mathcal{L}_{L1}^{2D} + \lambda_{giou}\mathcal{L}_{GIoU}\\ & \quad + \lambda_{mask}\mathcal{L}_{CE}^{mask} +   \lambda_{dice}\mathcal{L}_{DICE}^{mask} + \lambda_{3d}\mathcal{L}_{L1}^{3D}.
\end{split}
\end{equation}
Specifically, $\mathcal{L}_{cls}$ is the focal loss for classification. $\mathcal{L}_{L1}^{2D}$ and $\mathcal{L}_{GIoU}$ optimize 2D boxes. 2D masks are optimized via cross-entropy ($\mathcal{L}_{CE}^{mask}$) and DICE losses ($\mathcal{L}_{DICE}^{mask}$) using the spatial dot product between $\mathbf{c}_j$ and $\mathcal{M}_{mask}^{(4)}$. Lastly, $\mathcal{L}_{L1}^{3D}$ constrains absolute 3D positions against unprojected pseudo ground-truth centroids, firmly anchoring 2D features to physical 3D space.

\begin{figure*}[t]
\centering
\includegraphics[width=\textwidth]{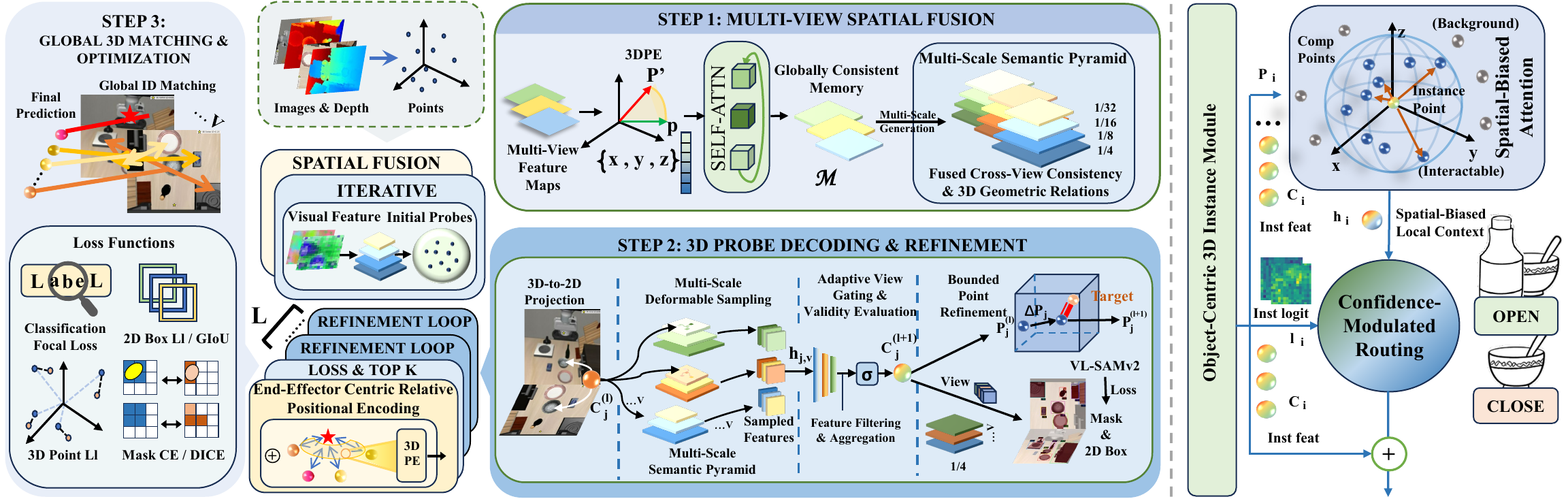}
\caption{
\textbf{Left}: Object-centric 3D instance modeling. 
\textbf{Right}: Spatially-conditioned geometry aggregation with uncertainty-guided routing.
}
\label{fig:overall_modules}
\end{figure*}

\subsection{Coordinate-Driven 3D Self-Supervised Predictor}
\label{subsec:ssl_predictor}

To address fragile reasoning under occlusion, we propose a \textbf{Coordinate-Driven 3D Self-Supervised Predictor}. We repurpose the masked self-supervised predictor to formulate 3D completion as a coordinate-conditioned feature-regression process for synthesizing unobserved structures.

\subsubsection{Target Coordinate Formulation and Geometric Completion}
\label{subsubsec:target_formulation}

We define Mask Coordinates ($\mathbf{P}_{mask}$) for self-supervised distillation and Completion Coordinates ($\mathbf{P}_{comp}$) for unobservable geometries. During training, asymmetric multi-view masking forces the network to leverage cross-view geometric priors instead of trivial spatial interpolation to infer missing structures, with unmasked regions forming the visible context $\mathcal{X}_{vis}$ anchored at $\mathbf{P}_{vis}$. To generate $\mathbf{P}_{comp}$, we back-project 2D masks into a raw point cloud, denoise it via dynamic density-based clustering, and use a pre-trained completion network~\cite{pointattn} to predict a dense shape. We isolate unobservable regions by sampling the farthest points from the observed geometry and then apply Farthest Point Sampling to obtain a uniform distribution. At inference, masking is disabled, and the predictor queries the continuous space directly with $\mathbf{P}_{comp}$.

\subsubsection{Position-View Conditioned Hybrid Attention}
\label{subsubsec:hybrid_attn}

Inferring unobserved features requires encoding both absolute coordinates and specific camera perspectives. We propose a Hybrid 3D Attention Mechanism that decouples identity injection from relative distance encoding. We inject a continuous Fourier 3D positional encoding $\Phi_{pos}(\cdot)$ and discrete camera view embedding $\mathbf{E}_{view}(\cdot)$ into the context $\mathcal{X}_{vis}$ and target queries $\mathcal{Z}_{tgt}$:
\begin{align}
\tilde{\mathcal{X}}_{vis} &= \mathcal{X}_{vis} + \Phi_{pos}(\mathbf{P}_{vis}) + \mathbf{E}_{view}(\mathbf{V}_{vis}), \\
\tilde{\mathcal{Z}}_{tgt} &= \mathcal{Z}_{tgt} + \Phi_{pos}(\mathbf{P}_{tgt}) + \mathbf{E}_{view}(\mathbf{V}_{tgt}),
\end{align}
where $\mathbf{V}_{vis}$ and $\mathbf{V}_{tgt}$ denote the respective view identifier vectors for the context and target coordinates, $\mathbf{P}_{tgt} = \mathbf{P}_{mask} \cup \mathbf{P}_{comp}$.
Continuous 3D RoPE is applied exclusively to the Query ($\mathbf{Q}$) and Key ($\mathbf{K}$) matrices to inject Euclidean distance priors. The core Hybrid Attention ($\operatorname{H-Attn}$) operator is defined as:
\begin{gather}
\mathbf{Q} = \operatorname{RoPE}_{3D}(\mathcal{X}_{q} \mathbf{W}_Q, \mathbf{P}_{q}), \\
\quad \mathbf{K} = \operatorname{RoPE}_{3D}(\mathcal{X}_{kv} \mathbf{W}_K, \mathbf{P}_{kv}), \\
\operatorname{H-Attn}(\mathcal{X}_{q}, \mathcal{X}_{kv}) = \operatorname{softmax}\Big( \frac{\mathbf{Q} \mathbf{K}^T}{\sqrt{d_k}} \Big) (\mathcal{X}_{kv} \mathbf{W}_V),
\end{gather}
where $d_k$ is the channel dimension per head. Each predictor block sequentially employs $\operatorname{H-Attn}$ for cross-attention ($\mathcal{X}_q = \tilde{\mathcal{Z}}_{tgt}, \mathcal{X}_{kv} = \tilde{\mathcal{X}}_{vis}$) to extract visual features from the unmasked context, and for self-attention ($\mathcal{X}_q = \mathcal{X}_{kv} = \tilde{\mathcal{Z}}_{tgt}$) to exchange spatial information. This sequential integration aligns structural consistencies in the final representations $\hat{\mathcal{Z}}_{tgt} = [\hat{\mathcal{Z}}_{mask}, \hat{\mathcal{Z}}_{comp}]$.

\subsubsection{Distillation Objective and Anti-Collapse Regularization}
\label{subsubsec:var_loss}

To stabilize continuous feature regression, representations are distilled from an EMA teacher network ($\theta_{\mathcal{T}} \leftarrow m \theta_{\mathcal{T}} + (1 - m) \theta_{\mathcal{S}}$). Continuous feature prediction is highly susceptible to minimizing reconstruction error by simply outputting an averaged vector. To prevent this, our objective governs the magnitude, direction, and variance of features:
\begin{equation}
\label{eq:ss_loss}
\mathcal{L}_{ss} = \lambda_{recon}\mathcal{L}_{recon} + \lambda_{cos}\mathcal{L}_{cos} + \lambda_{var}\mathcal{L}_{var}.
\end{equation}
Here, $\mathcal{L}_{recon}$ is a Smooth-L1 loss for magnitude alignment. $\mathcal{L}_{cos}$ is a cosine distance penalty aligning semantic angles independent of their norms:
\begin{equation}
\mathcal{L}_{cos} = 1 - \frac{1}{N_{mask}} \sum_{i=1}^{N_{mask}} \frac{\hat{\mathbf{z}}_i \cdot \mathbf{y}_i}{\|\hat{\mathbf{z}}_i\|_2 \|\mathbf{y}_i\|_2},
\end{equation}
where $\mathbf{y}_i \in \mathcal{Y}_{tea}$ is the corresponding teacher target token. Finally, $\mathcal{L}_{var}$ is a variance regularization loss to avoid spatial homogeneity:
\begin{equation}
\mathcal{L}_{var} = \frac{1}{D} \sum_{d=1}^D \operatorname{ReLU}\Big(\operatorname{std}_d(\mathcal{Y}_{tea}) - \operatorname{std}_d(\hat{\mathcal{Z}}_{mask})\Big),
\end{equation}
where $\operatorname{std}_d(\cdot)$ computes the standard deviation for the $d$-th channel.

\begin{table*}[t]
\centering
\caption{\textbf{Main Results on LIBERO-Plus.} We report success rates (\%) across 7 perturbation categories. \textbf{Bold} and \underline{underline} indicate the best and second-best results, respectively. \textbf{Ours} achieves the highest average success rate, significantly outperforming state-of-the-art baselines under severe visual and spatial variations.} 
\label{tab:main_results_libero}
\resizebox{0.85\textwidth}{!}{
\begin{tabular}{l| ccccccc| c<{\hspace{1.0em}}}
\toprule
\textbf{Methods} &  \textbf{Camera} & \textbf{Robot} & \textbf{Language} & \textbf{Light} & \textbf{Background} & \textbf{Noise} & \multicolumn{1}{c}{\textbf{Layout}} & \textbf{Avg.} \\
\hline
\gr OpenVLA~\cite{openvla}   & 0.8 & 3.5 & 23.0 & 8.1 & 34.8 & 15.2 & 28.5 & 15.6 \\
NORA~\cite{nora}   & 2.2 & 37.0 & 65.1 & 45.7 & 58.6 & 12.8 & 62.1 & 39.0 \\
\gr UniVLA~\cite{univla} & 1.8 & 46.2 & 69.6 & 69.0 & 81.0 & 21.2 & 31.9 & 42.9 \\
$\pi_0$-Fast~\cite{pi_0_fast} & 65.1 & 21.6 & 61.0 & 73.2 & 73.2 & 74.4 & 68.8 & 61.6 \\
\gr RIPT-VLA~\cite{riptvla} & 55.2 & 31.2 & 77.6 & 88.4 & 91.6 & 73.5 & 74.2 & 68.4 \\
OpenVLA-OFT~\cite{openvla_oft}  & 56.4 & 31.9 & 79.5 & 88.7 & 93.3 & 75.8 & 74.2 & 69.6 \\
\gr $\pi_0$~\cite{pi_0} & 61.0 & 40.8 & 63.5 & 89.3 & 84.1 & 80.1 & 76.4 & 69.4 \\
3DThinkVLA ~\cite{3dthinkvla} &73.8 &64.5 &78.0 &\textbf{98.4} &\underline{94.8} &84.7 &81.5 &81.0 \\
\gr SG-WAM~\cite{sgwam} &58.6	&48.9	&81.4	&89.8	
&86.1	&80.7	&74.2	&73.0 \\
GAM~\cite{gam}	&\textbf{83.1}	&70.0	&84.8	&97.2	&94.3	&\textbf{95.3}	&79.1	&\underline{85.5} \\
\hline
$\pi_{0.5}$~\cite{pi_05} & 71.2 & \underline{77.1} & \textbf{89.9} & 94.7 & 94.0 & 84.2 & \underline{84.3} & 84.2 \\

\good $\pi_{0.5}$+3DVLA
& \ppmark{\underline{75.6}}{+4.4}
& \ppmark{\textbf{77.4}}{+0.3}
& \ppmark{\underline{88.6}}{-1.3}
& \ppmark{\underline{97.4}}{+2.7}
& \ppmark{\textbf{97.7}}{+3.7}
& \ppmark{\underline{85.3}}{+1.1}
& \ppmark{\textbf{86.6}}{+2.3}
& \ppmark{\textbf{86.0}}{+1.8}\ppflush  \\
\noalign{\hrule height \heavyrulewidth}
\end{tabular}
}
\end{table*}

\begin{table}[t]
    \centering

    \caption{\textbf{Main Results on RoboTwin 2.0.}}
    \label{tab:design_adapter_ablation}
    \vspace{-3pt}

    \footnotesize
    \resizebox{0.6\linewidth}{!}{%
    \begin{tabular}{l|rr<{\hspace{1.0em}}}
        \toprule
        \multicolumn{1}{l}{\textbf{Method}} & \textbf{Easy} & \textbf{Hard} \\
        \hline
        DP~\cite{diffusion} & 28.0 & 0.6 \\
        \gr ACT~\cite{act} & 29.7 & 1.7 \\
        RDT~\cite{rdt} & 34.5 & 13.7 \\
        \gr DP3~\cite{dp3} & 55.2 & 5.0 \\
        \hline
        $\pi_0$~\cite{pi_0} & 46.4 & 16.3 \\
        \good \textbf{$\pi_0$+3DVLA}
        & \ppmark{54.5}{+8.1}
        & \ppmark{23.2}{+6.9}\ppflush \\
        \hline
        X-VLA~\cite{xvla} & 70.0 & 39.0 \\
        \good \textbf{X-VLA+3DVLA}
        & \ppmark{\textbf{80.6}}{+10.6}
        & \ppmark{\textbf{48.2}}{+9.2}\ppflush \\
        \noalign{\hrule height \heavyrulewidth}
    \end{tabular}%
    }

    \vspace{8pt}

    \caption{\textbf{Real-world Evaluation on Quantum-2 Robot.}
    We report the mean and standard deviation of success rates (\%)
    over 5 independent training runs. Avg. denotes the average
    success rate across tasks.}
    \label{tab:real_robot}
    \vspace{-3pt}

    \footnotesize
    \setlength{\tabcolsep}{3pt}
    \resizebox{\linewidth}{!}{%
    \begin{tabular}{l|c|ccc}
        \toprule
        \multicolumn{1}{c}{Method} & \multicolumn{1}{c}{Avg.} & Pick-and-Place & Fold Clothes & Press Button \\
        \hline
        $\pi_0.5$ & 62.5 & $68.4 \pm 7.6$ & $32.8 \pm 6.4$ & $86.2 \pm 2.6$ \\
        \good \textbf{3DVLA} & 70.3 &
        $\mathbf{74.0 \pm 4.6}$ &
        $\mathbf{50.4 \pm 5.1}$ &
        $\mathbf{86.6 \pm 2.5}$ \\
        \noalign{\hrule height \heavyrulewidth}
    \end{tabular}%
    }
\end{table}

\subsection{Spatially-Conditioned Geometry Aggregation}
\label{subsec:geometric_aggregation}

\subsubsection{End-Effector Centric Relative Positional Encoding}
\label{subsubsec:relative_encoding}

In downstream manipulation, actions are inherently executed relative to the robot's gripper. Providing only absolute world coordinates forces the policy to implicitly learn complex kinematic transformations. To bypass this, we explicitly ground the geometric representations in end-effector-centric spatial offsets: $\Delta \mathbf{p} = \mathbf{p}_{obj} - \mathbf{p}_{ee}$. To capture high-frequency spatial variations, $\Delta \mathbf{p}$ is expanded using sinusoidal functions across $F$ frequencies:
\begin{equation}
\begin{split}
\Phi_{freq}(\Delta \mathbf{p}) &= [\Delta \mathbf{p}, \sin(2^0 \pi \Delta \mathbf{p}), \\ &\quad \cos(2^0 \pi \Delta \mathbf{p}), \dots, \cos(2^{F-1} \pi \Delta \mathbf{p})].
\end{split}
\end{equation}
This continuous vector is projected via an MLP to match the token dimension. The resulting affordance embeddings are added to the corresponding instance and completion tokens. Finally, these spatially grounded tokens are concatenated with the visual tokens and fed into the action expert, thereby perfectly fulfilling the plug-and-play design without altering the base VLA architecture.

\subsubsection{Uncertainty-Guided Instance Geometric Routing}
\label{subsubsec:uncertainty_routing}

Fusing dense 3D completion features directly into the semantic states of instance probes ($\mathbf{c}_i$) risks corrupting highly confident visual representations. We propose Uncertainty-Guided Instance Geometric Routing to adaptively inject 3D geometry based on physical proximity and explicit detection uncertainty.

\textbf{Spatial-Biased Local Context.} Using 3D Euclidean distance, we retrieve the $K$-nearest completion points $\mathcal{N}(i)$ for the instance centered at $\mathbf{p}_i$. The local geometric context $\mathbf{h}_i$ is aggregated via cross-attention, where relative physical offsets dictate the attention bias:
\begin{equation}
\alpha_{ij} = \operatorname{softmax}_j \left( \frac{\mathbf{q}_i \mathbf{k}_j^T}{\sqrt{d_k}} + \operatorname{MLP}_{bias}(\mathbf{p}_j - \mathbf{p}_i) \right),
\end{equation}
\begin{equation}
\mathbf{h}_i = \sum_{j \in \mathcal{N}(i)} \alpha_{ij} \mathbf{v}_j \mathbf{W}_{out},
\end{equation}

where $\mathbf{q}_i$ is linearly projected from the instance semantic state $\mathbf{c}_i$, while $\mathbf{k}_j$ and $\mathbf{v}_j$ are projected from the local 3D completion feature at point $\mathbf{p}_j$. 
Crucially, the dot product $\mathbf{q}_i \mathbf{k}_j^T$ captures semantic consistency, while the penalty term $\operatorname{MLP}_{bias}(\mathbf{p}_j - \mathbf{p}_i)$ encodes spatial constraints. This joint formulation ensures that the instance only attends to geometries that are both semantically matched and physically adjacent.

\begin{figure*}[t]
\centering
\includegraphics[width=\textwidth]{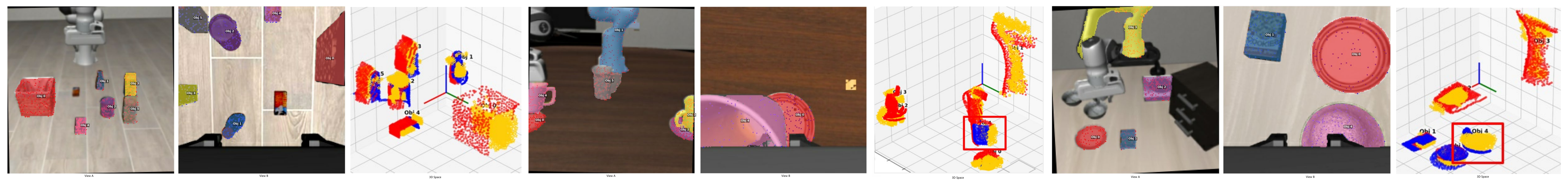}
\caption{\textbf{Multi-view instance consistency and 3D completion.} \textbf{Left \& Center:} Globally consistent masks and projected points generated from instance tokens across different views. \textbf{Right:} 3D geometry completion results in world space. Red and blue dots are visible points unprojected from respective views; yellow stars represent the predicted full geometry completed by our model.}
\label{fig:IS}
\end{figure*}

\begin{figure*}[t]
\centering
\includegraphics[width=\textwidth]{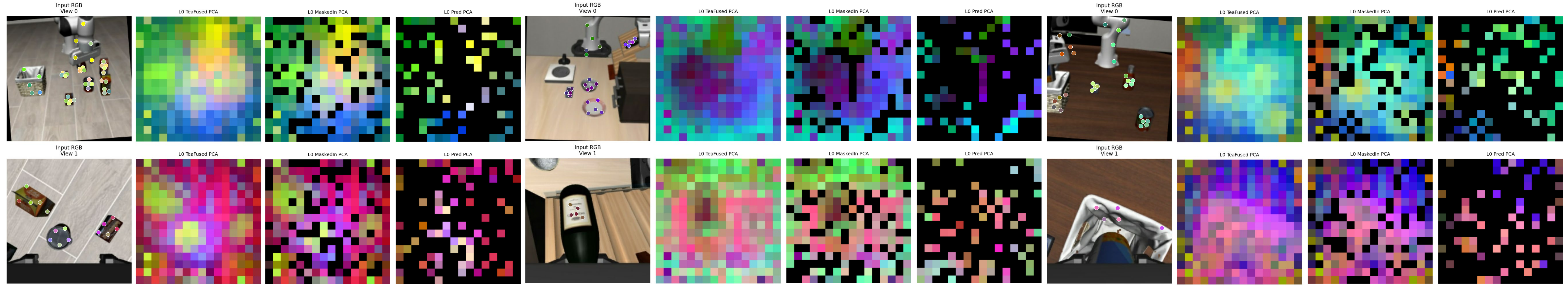}
\caption{\textbf{Coordinate-Driven 3D Self-Supervised Predictor.} Illustration of the predictor's geometry synthesis capabilities under severe occlusion and visual perturbations.}
\label{fig:predictor}
\vspace{-10pt}
\end{figure*}

\textbf{Confidence-Modulated Routing.} To ensure the network absorbs 3D context only when necessary, we quantify the ambiguity of the 2D observation. Using the classification logit $l_i$ generated by the instance module, we define an uncertainty score $u_i = 1 - \sigma(l_i) \in [0, 1]$, where $\sigma$ denotes the sigmoid function. We formulate a dynamic routing gate $\mathbf{g}_i \in \mathbb{R}^D$ conditioned on the instance feature, local 3D context, and uncertainty score:
\begin{equation}
\mathbf{g}_i = \sigma \Big( \tau \cdot \operatorname{MLP}_{gate}\big( [\mathbf{c}_i \,\|\, \mathbf{h}_i \,\|\, u_i] \big) \Big),
\end{equation}
where $[\cdot \,\|\, \cdot]$ denotes channel concatenation. Crucially, the final linear layer of $\operatorname{MLP}_{gate}$ is zero-initialized with a constant bias of $-3.0$. This guarantees an identity mapping ($\mathbf{g}_i \approx \mathbf{0}$) at the onset of training, protecting the pre-trained 2D features from gradient chaos. The instance representation is ultimately updated via $\hat{\mathbf{c}}_i = \mathbf{c}_i + \mathbf{g}_i \odot \operatorname{LayerNorm}(\mathbf{h}_i)$.

\section{Experiments}
\label{sec:experiments}

\subsection{Benchmarks and metrics}
\label{subsec:benchmarks}

We conduct experiments on two highly challenging simulation benchmarks. We strictly follow their official training splits and evaluation protocols.
\textbf{LIBERO-Plus} \cite{liberoplus} is a robotic benchmark built on LIBERO platform. It introduces seven distinct perturbation dimensions (e.g., Camera, Layout, Background, Light). 
\textbf{RoboTwin 2.0} \cite{robotwin} emphasizes physics-rich, contact-heavy manipulation requiring precise 3D geometric perception. It features diverse multi-view setups and severe self-occlusions across articulated, deformable, and long-horizon tasks.
These benchmarks use success rate as the evaluation metric.

\subsection{Main Results}
\label{subsec:main_results}

\textbf{Zero-Shot Transfer on LIBERO-Plus.} Under zero-shot transfer from LIBERO to LIBERO-Plus, 3DVLA achieves a new state-of-the-art average success rate of 86.0\% (Table \ref{tab:main_results_libero}). It is highly resilient to spatial shifts (e.g., \textit{Camera}, \textit{Layout}) and visual perturbations (e.g., \textit{Light}, \textit{Background}). We attribute these gains to anchoring features in continuous 3D space for multi-view consistency and to the coordinate-driven predictor that reconstructs occluded task-relevant geometry. We dissect these mechanisms in Section \ref{subsec:ablations}.

\textbf{Comprehensive Manipulation on RoboTwin 2.0.} We further benchmark 3DVLA on RoboTwin 2.0 to assess physics-rich interaction. Demonstrating plug-and-play compatibility, we integrate 3DVLA into $\pi_0$ and X-VLA, and yield consistent and substantial improvements on both Easy and Hard settings (Table \ref{tab:design_adapter_ablation}). This confirms the effectiveness of our proposed method, while 3DVLA adds only $\sim$0.086B parameters and 8.6\% (21.22 ms on an A100) computational overhead.

\textbf{Real-world Evaluation.}
To evaluate the applicability of 3DVLA under real-world sensing and control, we conduct experiments on a Quantum-2 robot.
For each task, we collect 100 teleoperated demonstrations and train five independent IL policies with different random seeds.
Table~\ref{tab:real_robot} reports the mean success rate and standard deviation over 50 test episodes.
Comparisons of baseline methods and 3DVLA, together with a real-world demonstration, are provided in the Appendix and Supplementary Video.

\begin{table*}[t]
\centering
\caption{\textbf{Overall Ablation Studies on LIBERO-Plus.} 
\textit{MSF}: Multi-View Spatial Fusion; {Inst.}: 3D Instance Module; \textit{Pred.}: 3D Self-Supervised \& Predictor; \textit{Rout.}: Uncertainty-Guided Routing for the Aggregation.
}
\label{tab:ablations}
\resizebox{\textwidth}{!}{
\begin{tabular}{l |cccc |ccccccc |c}
\toprule
\multicolumn{1}{l}{\textbf{Model}} &  \textbf{MSF} &\textbf{Inst.} & \textbf{Pred.} & \multicolumn{1}{c}{\textbf{Rout.}}& \textbf{Camera} & \textbf{Robot} & \textbf{Language} & \textbf{Light} & \textbf{Background} & \textbf{Noise} & \multicolumn{1}{c}{\textbf{Layout}} & \textbf{Avg.} \\
\hline
\gr Baseline & & & & & 71.2 & 77.1 & \textbf{89.9} & 94.7 & 94.0 & 84.2 & 84.3 & 84.2 \\
MSF only & \checkmark & & & & 71.2 & 77.5 & 89.0 & 95.1 & 94.1 & 85.0 & 84.6 & 84.5 \\
\gr \quad A & \checkmark & \checkmark & & & 71.4 & 78.9 & 87.9 & 96.0 & 94.4 & 86.5 & 86.3 & 85.0 \\
\quad B & \checkmark & \checkmark & \checkmark & & 73.6 & \textbf{79.4} & 88.2 & 97.1 & 96.5 & 85.2 & 86.4 & 85.7 \\
\good \quad \textbf{C (Ours)}& \checkmark & \checkmark & \checkmark & \checkmark & \textbf{75.6} & 77.4 & 88.6 & \textbf{97.4} & \textbf{97.7} & \textbf{85.3} & \textbf{86.6} & \textbf{86.0} \\
\noalign{\hrule height \heavyrulewidth}
\end{tabular}
}
\end{table*}


\begin{table}[t]
    \centering

    \caption{\textbf{Ablation of the 3D Instance Module.}}
    \label{tab:ablation_instance}
    \vspace{-3pt}
    \footnotesize
    \resizebox{0.9\linewidth}{!}{%
    \begin{tabular}{l|cc}
        \toprule
        \multicolumn{1}{c}{\textbf{Variant}} &
        \textbf{Layout} & \textbf{Avg.} \\
        \hline
        w/o Spatial Fusion & 84.6 & 84.6 \\
        \gr 2D Matching (vs. Global 3D) & 85.4 & 84.5 \\
        w/o End-Effector Relative PE & 86.0 & 84.8 \\
        \good \textbf{Full Instance Module (A)} &
        \textbf{86.3} & \textbf{85.0} \\
        \noalign{\hrule height \heavyrulewidth}
    \end{tabular}%
    }

    \vspace{8pt}

    \caption{\textbf{Comprehensive Ablation Studies on RoboTwin 2.0.}}
    \label{tab:full_ablation}
    \vspace{-3pt}
    \footnotesize
    \resizebox{0.8\linewidth}{!}{%
    \begin{tabular}{c|ccc|cc}
        \toprule
        \multicolumn{1}{c}{\textbf{Model}} & 
        \textbf{Inst.} & \textbf{Pred.} &
        \multicolumn{1}{c}{\textbf{Rout.}}
        & \textbf{Easy} & \textbf{Hard} \\
        \hline
        $\pi_0$ & & & & 46.4 & 16.3 \\
        \gr A & \checkmark & & & 50.1 & 18.9 \\
        B & \checkmark & \checkmark & & 53.4 & 21.8 \\
        \good \textbf{C (Ours)} &
        \checkmark & \checkmark & \checkmark &
        \textbf{54.5} & \textbf{23.2} \\
        \hline
        X-VLA & & & & 70.0 & 39.0 \\
        \gr A & \checkmark & & & 72.3 & 41.7 \\
        B & \checkmark & \checkmark & & 78.5 & 46.5 \\
        \good \textbf{C (Ours)} &
        \checkmark & \checkmark & \checkmark &
        \textbf{80.6} & \textbf{48.2} \\
        \noalign{\hrule height \heavyrulewidth}
    \end{tabular}%
    }
\end{table}


\subsection{Ablation Study}
\label{subsec:ablations}

We examine the contribution of each proposed module via systematic ablation experiments. To provide a comprehensive view, we analyze the progressive macro-integration across both RoboTwin 2.0 (Table \ref{tab:full_ablation}) and LIBERO-Plus (Table \ref{tab:ablations}). We then delve into micro-ablations detailing specific architectural choices (Table \ref{tab:ablation_instance}).

\textbf{3D Instance Module \& Spatial Fusion.} 
We first isolate the contribution of Multi-View Spatial Fusion (MSF).
Adding MSF alone improves the overall success rate from 84.2\% to
84.5\%, indicating that shared 3D feature fusion provides a modest
improvement by itself.  The 3D Instance Module (A) raises the average success rate on both LIBERO-Plus and RoboTwin 2.0, confirming that direct 3D entity extraction overcomes flat 2D region reasoning. 
1) Removing voxel-free spatial fusion drops performance from 85.0\% to 84.6\%, showing that pre-decoding multi-view consistency is necessary to align views into coherent memory (Figure \ref{fig:IS}, Left \& Center).
2) Replacing our global 3D bipartite matching with heuristic 2D matching drops the average to 84.5\%, as 2D matching inevitably reintroduces cross-view ID conflicts when cameras shift. 
3) Omitting the end-effector relative positional encoding degrades performance (84.8\%). Logically, action execution is intrinsically relative to the robot's gripper. Feeding absolute world coordinates forces the policy to implicitly learn complex kinematic transformations, whereas our relative encoding provides a direct, actionable spatial offset.

\textbf{Coordinate-Driven 3D Self-Supervised Predictor.} 
Introducing the 3D Self-Supervised Predictor (B) directly targets fragile reasoning under occlusion. This addition increases the LIBERO-Plus average to 85.7\% and yields a substantial jump in RoboTwin 2.0's \textit{Hard} tasks (18.9\% $\rightarrow$ 21.8\%). Specifically, as visualized in Figure \ref{fig:predictor}, the significant gain in the \textit{Camera} split (71.4\% $\rightarrow$ 73.6\%) confirms that actively inferring unseen topologies prevents the policy from failing when task-relevant parts are hidden.


\textbf{Uncertainty-Guided Instance Geometric Routing.}
Finally, we evaluate the Uncertainty-Guided Instance Geometric Routing (C).
Model B feeds completion and instance tokens separately to the action expert.
Model C instead integrates local 3D geometry into instance representations based on observation uncertainty.
This adaptive fusion improves LIBERO-Plus performance by +2.0\% under \textit{Camera} variations and achieves the best results on both LIBERO-Plus (86.0\% average) and RoboTwin 2.0 (48.2\% Hard).
The improvement comes from two aspects: 3D geometry fusion reduces the burden of learning spatial-semantic alignment, while uncertainty-aware routing prevents unreliable synthesized features from contaminating confident visual representations.

\section{Conclusion}
\label{sec:conclusion}

In this paper, we propose 3DVLA, a plug-and-play framework that enriches VLAs with robust 3D reasoning while remaining fully compatible with VLM pre-training and requiring no additional manual annotations. Our method integrates pervasive multi-view consistency constraints for viewpoint-invariant representations, instance tokens for 3D instance awareness, and a masked, self-supervised 3D-encoding branch with a retained predictor for visual token completion. Experiments on the LIBERO-Plus and RoboTwin 2.0 benchmarks demonstrate that 3DVLA consistently improves manipulation performance over multiple baselines, validating both its effectiveness and its practical plug-and-play compatibility.

\clearpage


{
\small
\bibliographystyle{ieeenat_fullname}
\bibliography{main}

\begin{thebibliography}{33}
\providecommand{\natexlab}[1]{#1}
\providecommand{\url}[1]{\texttt{#1}}
\expandafter\ifx\csname urlstyle\endcsname\relax
  \providecommand{\doi}[1]{doi: #1}\else
  \providecommand{\doi}{doi: \begingroup \urlstyle{rm}\Url}\fi

\bibitem[Assran et~al.(2023)Assran, Duval, Misra, Bojanowski, Vincent, Rabbat, LeCun, and Ballas]{ijepa}
Mahmoud Assran, Quentin Duval, Ishan Misra, Piotr Bojanowski, Pascal Vincent, Michael Rabbat, Yann LeCun, and Nicolas Ballas.
\newblock Self-supervised learning from images with a joint-embedding predictive architecture.
\newblock In \emph{CVPR}, 2023.

\bibitem[Black et~al.(2024)Black, Brown, Driess, Esmail, Equi, Finn, Fusai, Groom, Hausman, Ichter, et~al.]{pi_0}
Kevin Black, Noah Brown, Danny Driess, Adnan Esmail, Michael Equi, Chelsea Finn, Niccolo Fusai, Lachy Groom, Karol Hausman, Brian Ichter, et~al.
\newblock $\pi_{0}$: A vision-language-action flow model for general robot control.
\newblock \emph{arXiv preprint arXiv:2410.24164}, 2024.

\bibitem[Brohan et~al.(2022)Brohan, Brown, Carbajal, Chebotar, Dabis, Finn, Gopalakrishnan, Hausman, Herzog, Hsu, et~al.]{rt1}
Anthony Brohan, Noah Brown, Justice Carbajal, Yevgen Chebotar, Joseph Dabis, Chelsea Finn, Keerthana Gopalakrishnan, Karol Hausman, Alex Herzog, Jasmine Hsu, et~al.
\newblock Rt-1: Robotics transformer for real-world control at scale.
\newblock In \emph{RSS}, 2022.

\bibitem[Bu et~al.(2025)Bu, Yang, Cai, Gao, Ren, Yao, Luo, and Li]{univla}
Qingwen Bu, Yanting Yang, Jisong Cai, Shenyuan Gao, Guanghui Ren, Maoqing Yao, Ping Luo, and Hongyang Li.
\newblock Learning to act anywhere with task-centric latent actions.
\newblock In \emph{RSS}, 2025.

\bibitem[Cen et~al.(2025)Cen, Yu, Yuan, Jiang, Huang, Guo, Li, Song, Luo, Wang, et~al.]{worldvla}
Jun Cen, Chaohui Yu, Hangjie Yuan, Yuming Jiang, Siteng Huang, Jiayan Guo, Xin Li, Yibing Song, Hao Luo, Fan Wang, et~al.
\newblock Worldvla: Towards autoregressive action world model.
\newblock \emph{arXiv preprint arXiv:2506.21539}, 2025.

\bibitem[Chen et~al.(2025)Chen, Chen, Chen, Cai, Liu, Li, Liang, Lin, Ge, Gu, et~al.]{robotwin}
Tianxing Chen, Zanxin Chen, Baijun Chen, Zijian Cai, Yibin Liu, Zixuan Li, Qiwei Liang, Xianliang Lin, Yiheng Ge, Zhenyu Gu, et~al.
\newblock Robotwin 2.0: A scalable data generator and benchmark with strong domain randomization for robust bimanual robotic manipulation.
\newblock \emph{arXiv preprint arXiv:2506.18088}, 2025.

\bibitem[Chi et~al.(2025)Chi, Xu, Feng, Cousineau, Du, Burchfiel, Tedrake, and Song]{diffusion}
Cheng Chi, Zhenjia Xu, Siyuan Feng, Eric Cousineau, Yilun Du, Benjamin Burchfiel, Russ Tedrake, and Shuran Song.
\newblock Diffusion policy: Visuomotor policy learning via action diffusion.
\newblock \emph{IJRR}, 2025.

\bibitem[Edwards et~al.(2019)Edwards, Sahni, Schroecker, and Isbell]{ilpo}
Ashley Edwards, Himanshu Sahni, Yannick Schroecker, and Charles Isbell.
\newblock Imitating latent policies from observation.
\newblock In \emph{ICML}, 2019.

\bibitem[Fei et~al.(2025)Fei, Wang, Shi, Dai, Cai, Qian, Ji, He, Zhang, Fei, et~al.]{liberoplus}
Senyu Fei, Siyin Wang, Junhao Shi, Zihao Dai, Jikun Cai, Pengfang Qian, Li Ji, Xinzhe He, Shiduo Zhang, Zhaoye Fei, et~al.
\newblock Libero-plus: In-depth robustness analysis of vision-language-action models.
\newblock \emph{arXiv preprint arXiv:2510.13626}, 2025.

\bibitem[Han et~al.(2026)Han, Jeon, Jung, Zurbr{\"u}gg, An, Portela, Hutter, Pollefeys, Kim, and Hong]{gam}
Jisang Han, Seonghu Jeon, Jaewoo Jung, Ren{\'e} Zurbr{\"u}gg, Honggyu An, Tifanny Portela, Marco Hutter, Marc Pollefeys, Seungryong Kim, and Sunghwan Hong.
\newblock Geometric action model for robot policy learning.
\newblock \emph{arXiv preprint arXiv:2606.17046}, 2026.

\bibitem[Hung et~al.(2025)Hung, Sun, Hong, Zadeh, Li, Tan, Majumder, Poria, et~al.]{nora}
Chia-Yu Hung, Qi Sun, Pengfei Hong, Amir Zadeh, Chuan Li, U Tan, Navonil Majumder, Soujanya Poria, et~al.
\newblock Nora: A small open-sourced generalist vision language action model for embodied tasks.
\newblock \emph{arXiv preprint arXiv:2504.19854}, 2025.

\bibitem[Intelligence et~al.(2025)Intelligence, Black, Brown, Darpinian, Dhabalia, Driess, Esmail, Equi, Finn, Fusai, et~al.]{pi_05}
Physical Intelligence, Kevin Black, Noah Brown, James Darpinian, Karan Dhabalia, Danny Driess, Adnan Esmail, Michael Equi, Chelsea Finn, Niccolo Fusai, et~al.
\newblock $\pi_{0.5}$: a vision-language-action model with open-world generalization.
\newblock \emph{arXiv preprint arXiv:2504.16054}, 2025.

\bibitem[Kim et~al.(2025{\natexlab{a}})Kim, Finn, and Liang]{openvla_oft}
Moo~Jin Kim, Chelsea Finn, and Percy Liang.
\newblock Fine-tuning vision-language-action models: Optimizing speed and success.
\newblock \emph{arXiv preprint arXiv:2502.19645}, 2025{\natexlab{a}}.

\bibitem[Kim et~al.(2025{\natexlab{b}})Kim, Pertsch, Karamcheti, Xiao, Balakrishna, Nair, Rafailov, Foster, Sanketi, Vuong, et~al.]{openvla}
Moo~Jin Kim, Karl Pertsch, Siddharth Karamcheti, Ted Xiao, Ashwin Balakrishna, Suraj Nair, Rafael Rafailov, Ethan~P Foster, Pannag~R Sanketi, Quan Vuong, et~al.
\newblock Openvla: An open-source vision-language-action model.
\newblock In \emph{CoRL}, 2025{\natexlab{b}}.

\bibitem[Lin and Wang(2025)]{vlsam2}
Zhiwei Lin and Yongtao Wang.
\newblock Vl-sam-v2: Open-world object detection with general and specific query fusion.
\newblock In \emph{NeurIPS}, 2025.

\bibitem[Liu et~al.(2025)Liu, Wu, Li, Tan, Chen, Wang, Xu, Su, and Zhu]{rdt}
Songming Liu, Lingxuan Wu, Bangguo Li, Hengkai Tan, Huayu Chen, Zhengyi Wang, Ke Xu, Hang Su, and Jun Zhu.
\newblock Rdt-1b: a diffusion foundation model for bimanual manipulation.
\newblock In \emph{ICLR}, 2025.

\bibitem[Lynch et~al.(2020)Lynch, Khansari, Xiao, Kumar, Tompson, Levine, and Sermanet]{lapa}
Corey Lynch, Mohi Khansari, Ted Xiao, Vikash Kumar, Jonathan Tompson, Sergey Levine, and Pierre Sermanet.
\newblock Learning latent plans from play.
\newblock In \emph{CoRL}, 2020.

\bibitem[Oquab et~al.(2023)Oquab, Darcet, Moutakanni, Vo, Szafraniec, Khalidov, Fernandez, Haziza, Massa, El-Nouby, et~al.]{dinov2}
Maxime Oquab, Timoth{\'e}e Darcet, Th{\'e}o Moutakanni, Huy Vo, Marc Szafraniec, Vasil Khalidov, Pierre Fernandez, Daniel Haziza, Francisco Massa, Alaaeldin El-Nouby, et~al.
\newblock Dinov2: Learning robust visual features without supervision.
\newblock \emph{TMLR}, 2023.

\bibitem[Pertsch et~al.(2025)Pertsch, Stachowicz, Ichter, Driess, Nair, Vuong, Mees, Finn, and Levine]{pi_0_fast}
Karl Pertsch, Kyle Stachowicz, Brian Ichter, Danny Driess, Suraj Nair, Quan Vuong, Oier Mees, Chelsea Finn, and Sergey Levine.
\newblock Fast: Efficient action tokenization for vision-language-action models.
\newblock \emph{arXiv preprint arXiv:2501.09747}, 2025.

\bibitem[Rukhovich et~al.(2022)Rukhovich, Vorontsova, and Konushin]{imvoxelnet}
Danila Rukhovich, Anna Vorontsova, and Anton Konushin.
\newblock Imvoxelnet: Image to voxels projection for monocular and multi-view general-purpose 3d object detection.
\newblock In \emph{WACV}, 2022.

\bibitem[Shi et~al.(2026)Shi, Zhang, Zhu, Li, Zhu, and Yuan]{3dthinkvla}
Jiaxin Shi, Xidong Zhang, Fucai Zhu, Zhe Li, Siyu Zhu, and Weihao Yuan.
\newblock 3dthinkvla: Endowing vision-language-action models with latent 3d priors via 3d-thinking-guided co-training.
\newblock \emph{arXiv preprint arXiv:2606.04436}, 2026.

\bibitem[Sun et~al.(2026)Sun, Zhang, Qi, Ren, Liu, Zhu, Sun, Jin, and Chen]{vlajepa}
Jingwen Sun, Wenyao Zhang, Zekun Qi, Shaojie Ren, Zezhi Liu, Hanxin Zhu, Guangzhong Sun, Xin Jin, and Zhibo Chen.
\newblock Vla-jepa: Enhancing vision-language-action model with latent world model.
\newblock In \emph{ECCV}, 2026.

\bibitem[Tan et~al.(2025)Tan, Dou, Zhao, and Kr{\"a}henb{\"u}hl]{riptvla}
Shuhan Tan, Kairan Dou, Yue Zhao, and Philipp Kr{\"a}henb{\"u}hl.
\newblock Interactive post-training for vision-language-action models.
\newblock \emph{arXiv preprint arXiv:2505.17016}, 2025.

\bibitem[Team et~al.(2024)Team, Ghosh, Walke, Pertsch, Black, Mees, Dasari, Hejna, Kreiman, Xu, et~al.]{octo}
Octo~Model Team, Dibya Ghosh, Homer Walke, Karl Pertsch, Kevin Black, Oier Mees, Sudeep Dasari, Joey Hejna, Tobias Kreiman, Charles Xu, et~al.
\newblock Octo: An open-source generalist robot policy.
\newblock In \emph{RSS}, 2024.

\bibitem[Wang et~al.(2024)Wang, Cui, Guo, Li, Liu, and Shen]{pointattn}
Jun Wang, Ying Cui, Dongyan Guo, Junxia Li, Qingshan Liu, and Chunhua Shen.
\newblock Pointattn: You only need attention for point cloud completion.
\newblock In \emph{AAAI}, 2024.

\bibitem[Xia et~al.(2025)Xia, Li, Lin, Wang, Wang, and Yang]{openad}
Zhongyu Xia, Jishuo Li, Zhiwei Lin, Xinhao Wang, Yongtao Wang, and Ming-Hsuan Yang.
\newblock Openad: Open-world autonomous driving benchmark for 3d object detection.
\newblock In \emph{NeurIPS}, 2025.

\bibitem[Xia et~al.(2026{\natexlab{a}})Xia, Lin, Wang, and Yang]{henet++}
Zhongyu Xia, Zhiwei Lin, Yongtao Wang, and Ming-Hsuan Yang.
\newblock Henet++: Hybrid encoding and multi-task learning for 3d perception and end-to-end autonomous driving.
\newblock \emph{IJCV}, 2026{\natexlab{a}}.

\bibitem[Xia et~al.(2026{\natexlab{b}})Xia, Tang, Wang, Wang, and Qin]{r4det}
Zhongyu Xia, Yousen Tang, Yongtao Wang, Zhifeng Wang, and Weijun Qin.
\newblock R4det: 4d radar-camera fusion for high-performance 3d object detection.
\newblock In \emph{CVPR}, 2026{\natexlab{b}}.

\bibitem[Ze et~al.(2024)Ze, Zhang, Zhang, Hu, Wang, and Xu]{dp3}
Yanjie Ze, Gu Zhang, Kangning Zhang, Chenyuan Hu, Muhan Wang, and Huazhe Xu.
\newblock 3d diffusion policy: Generalizable visuomotor policy learning via simple 3d representations.
\newblock In \emph{RSS}, 2024.

\bibitem[Zhao et~al.(2026)Zhao, Zhang, Su, Wang, Li, Yang, Tay, Ang~Jr, and Zhu]{sgwam}
Ruiteng Zhao, Zhengshen Zhang, Yue Su, Wenshuo Wang, Jiahui Li, Zhiyuan Yang, Francis~EH Tay, Marcelo~H Ang~Jr, and Haiyue Zhu.
\newblock Sg-wam: Self-guided world modeling in geometry-aware policy space.
\newblock \emph{arXiv preprint arXiv:2608.01397}, 2026.

\bibitem[Zhao et~al.(2023)Zhao, Kumar, Levine, and Finn]{act}
Tony~Z Zhao, Vikash Kumar, Sergey Levine, and Chelsea Finn.
\newblock Learning fine-grained bimanual manipulation with low-cost hardware.
\newblock In \emph{RSS}, 2023.

\bibitem[Zheng et~al.(2026)Zheng, Li, Wang, Liu, Kang, Feng, Zheng, Zou, Chen, Zeng, et~al.]{xvla}
Jinliang Zheng, Jianxiong Li, Zhihao Wang, Dongxiu Liu, Xirui Kang, Yuchun Feng, Yinan Zheng, Jiayin Zou, Yilun Chen, Jia Zeng, et~al.
\newblock X-vla: Soft-prompted transformer as scalable cross-embodiment vision-language-action model.
\newblock In \emph{ICLR}, 2026.

\bibitem[Zitkovich et~al.(2023)Zitkovich, Yu, Xu, Xu, Xiao, Xia, Wu, Wohlhart, Welker, Wahid, et~al.]{rt2}
Brianna Zitkovich, Tianhe Yu, Sichun Xu, Peng Xu, Ted Xiao, Fei Xia, Jialin Wu, Paul Wohlhart, Stefan Welker, Ayzaan Wahid, et~al.
\newblock Rt-2: Vision-language-action models transfer web knowledge to robotic control.
\newblock In \emph{CoRL}, 2023.

\end{thebibliography}
}

\clearpage
\appendix

\section{Implementation Details}
\label{sec:implementation_details}

To facilitate future research and ensure full reproducibility, we will \textit{open-source} our complete codebase, pre-trained model weights, and evaluation scripts upon publication. Below, we detail the architectural hyperparameters, optimization strategies, and computational costs associated with 3DVLA.

\subsection{Architectural Hyper-parameters}
A core advantage of 3DVLA is its parameter efficiency. During training, the large underlying Vision-Language Model (VLM) backbone is kept entirely frozen, while only the newly introduced 3D perception modules and the downstream action expert are actively updated.
For the \textit{Object-Centric 3D Instance Module}, we initialize $N_q = 300$ state probes to ensure comprehensive scene coverage. Through matching and uncertainty filtering, these probes are adaptively distilled into a maximum of 16 highly confident instance tokens per scene. 
For the \textit{Coordinate-Driven 3D Predictor}, to balance computational efficiency and geometric fidelity, the network processes at most 6 occluded object instances per observation, with each instance uniformly sampled to ensure it contains exactly 5 completion coordinate queries.

\subsection{Two-Stage Optimization Procedure}
We emphasize that this optimization is performed during the post-training stage after loading publicly available pretrained VLA checkpoints. To ensure stable optimization and prevent continuous feature regression from dominating or collapsing the early learning phase, we decouple 3DVLA's training into a progressive two-stage pipeline.

\textbf{Stage 1: Spatial Fusion and Instance Extraction.} 
In the first stage, we focus exclusively on learning view-consistent physical entities. We jointly train the Multi-View Spatial Fusion module, the 3D Instance Module, and the action expert. This stage is trained for 5,000 steps with an effective batch size of 64.

\textbf{Stage 2: 3D Completion and Geometry Aggregation.}
In the second stage, we freeze the weights of the modules trained in Stage 1 to preserve the stable 2D-to-3D geometric anchoring. We then exclusively train the 3D Self-Supervised Predictor, the Geometry Aggregation module, and continue updating the action expert. This phase is also trained for 5,000 steps with a batch size of 64.

\subsection{Optimization and Loss Coefficients}
The network is optimized using the AdamW optimizer with a gradient clipping norm of 0.5 and a weight decay of $1 \times 10^{-4}$. We employ a cosine learning rate decay schedule, featuring 1,200 warmup steps to reach a peak learning rate of $3 \times 10^{-6}$, followed by 3,800 decay steps down to $1 \times 10^{-8}$.

For the joint geometric alignment loss (Equation~\ref{eq:joint_loss}), the balancing coefficients are empirically set to precisely match the scale of heterogeneous geometric properties: $\lambda_{cls} = 2.0$, $\lambda_{box} = 5.0$, $\lambda_{giou} = 2.0$, $\lambda_{mask} = 5.0$, $\lambda_{dice} = 5.0$, and $\lambda_{3d} = 5.0$. 
For the self-supervised distillation objective (Equation~\ref{eq:ss_loss}), to strictly prevent feature homogeneity, we strongly penalize variance collapse while maintaining directional alignment: $\lambda_{recon} = 1.0$, $\lambda_{cos} = 0.1$, and $\lambda_{var} = 10.0$. The Exponential Moving Average (EMA) momentum for updating the teacher network is set to $m = 0.999$.

\subsection{Computational Cost}
The entire two-stage training process is highly efficient and is conducted on a single compute node equipped with 4 NVIDIA A800 GPUs. Stage 1 takes approximately 5 hours, and Stage 2 takes roughly 9 hours, bounding the total training pipeline to merely 14 hours. 

During evaluation, due to the physics-rich nature of the environments and the requirement for comprehensive validation across diverse perturbation setups, the testing phase is heavily bottlenecked by simulation rendering and physical stepping rather than model inference. We scale the evaluation across 8 NVIDIA A800 GPUs, which take approximately 14 hours to complete the extensive simulation benchmark suites.

\section{Additional Experimental Analysis}
\label{sec:additional_analysis}

\subsection{Ablation on Self-Supervised Masking Ratio}
\label{sec:mask_ablation}

As shown In Table~\ref{tab:masking_ratio}, we evaluate the masking ratio during self-supervised training. A lower ratio of 25\% yields sub-optimal results (85.5\%) because the network bypasses 3D reconstruction by simply interpolating features from adjacent visible 2D pixels. Conversely, our 50\% asymmetric masking explicitly forces the predictor to perform true cross-view spatial reasoning, yielding the highest structural fidelity.

\begin{table}[h]
    \centering
    \caption{Ablation on the masking ratio for the self-supervised training on LIBERO-Plus.}
    \label{tab:masking_ratio}
    \renewcommand{\arraystretch}{1.2}
    \begin{tabular}{l|ccc}
        \toprule
        Masking Ratio & Camera & Noise & Avg. \\
        \midrule
        25\% Masking & 73.1 & 85.0 & 85.5 \\
        \good \textbf{50\% Masking (B)} & \textbf{73.6} & \textbf{85.2} & \textbf{85.7} \\
        \bottomrule
    \end{tabular}
\end{table}

\begin{table*}[h]
    \centering
    \caption{Robustness evaluation under different noise perturbations. All perturbations are applied only during inference, where $a$ denotes the artificially added noise ratio.}
    \label{tab:noise_robustness}
    \resizebox{0.6\textwidth}{!}{
        \begin{tabular}{l cccc cccc c}
            \toprule
            & \multicolumn{4}{c}{Instance Noise} & \multicolumn{4}{c}{Completion Noise} \\
            \cmidrule(lr){2-5} \cmidrule(lr){6-9}
            \multicolumn{1}{c}{$a$} & 0 & 0.3 & 0.5 & 1 & 0 & 0.3 & 0.5 & 1 \\
            
            \midrule
            \multicolumn{10}{c}{\textbf{LIBERO-Plus}} \\
            \midrule
            Baseline Pi0.5 & 84.2 & - & - & - & 84.2 & - & - & - \\
            \good \textbf{Pi0.5 + 3DVLA} & \textbf{86.0} & \textbf{85.4} & \textbf{85.7} & 84.1 & \textbf{86.0} & \textbf{85.6} & \textbf{85.1} & \textbf{85.0} \\
            
            \midrule
            \multicolumn{10}{c}{\textbf{RoboTwin 2.0}} \\
            \midrule
            Baseline Pi0 & 46.4 & - & - & - & 46.4 & - & - & - \\
            \good \textbf{Pi0 + 3DVLA} & \textbf{54.5} & \textbf{53.3} & \textbf{50.8} & \textbf{46.9} & \textbf{54.5} & \textbf{53.9} & \textbf{51.6} & \textbf{50.1} \\
            \bottomrule
        \end{tabular}
    }
\end{table*}

\begin{table*}[h]
    \centering
    \caption{Multi-seed evaluation on LIBERO-Plus. Results are reported as Mean $\pm$ Standard Deviation (\%) over five independent runs.}
    \label{tab:seed_libero}
    \renewcommand{\arraystretch}{1.2} 
    \resizebox{\textwidth}{!}{
        \begin{tabular}{l ccccccc c} 
            \toprule
            Model & Camera & Robot & Language & Light & Background & Noise & Layout & Avg \\
            \midrule
            Baseline Pi0.5         & 70.9$\pm$2.1          & 76.8$\pm$1.8          & 89.5$\pm$1.5          & 94.4$\pm$1.2          & 93.7$\pm$1.6          & 84.5$\pm$2.0          & 83.9$\pm$2.2          & 83.9$\pm$1.2 \\
            \good \textbf{Pi0.5 + 3DVLA} & \textbf{75.3$\pm$1.4} & \textbf{77.8$\pm$1.5} & 88.9$\pm$1.6          & \textbf{97.2$\pm$0.9} & \textbf{96.9$\pm$0.8} & \textbf{85.7$\pm$1.7} & \textbf{87.0$\pm$1.5} & \textbf{86.2$\pm$0.8} \\
            \bottomrule
        \end{tabular}
    }
\end{table*}

\begin{table}[h]
    \centering
    \caption{Multi-seed evaluation on RoboTwin 2.0. Results are reported as Mean $\pm$ Standard Deviation (\%) over five independent runs.}
    \label{tab:seed_robotwin}
    \renewcommand{\arraystretch}{1.2}
    \begin{tabular}{l cc} 
        \toprule
        Model & Easy & Hard \\
        \midrule
        Baseline Pi0           & 46.2$\pm$1.7          & 16.6$\pm$2.9 \\
        \good \textbf{Pi0 + 3DVLA}   & \textbf{54.8$\pm$1.4} & \textbf{23.1$\pm$2.2} \\
        \midrule
        Baseline X-VLA         & 69.7$\pm$0.9          & 39.0$\pm$3.1 \\
        \good \textbf{X-VLA + 3DVLA} & \textbf{80.6$\pm$1.2} & \textbf{48.2$\pm$2.4} \\
        \bottomrule
    \end{tabular}
\end{table}

\subsection{Robustness Analysis}
\label{sec:robustness}

Existing VLA models have rarely explored 3D understanding, largely because manual 3D annotation is prohibitively expensive and it is difficult to effectively integrate such 3D information into the decision-making process. We believe that one of the key advantages of 3DVLA is its ability to fully leverage the prior information from existing models. The modules we added are not a simple stack. Instead, they are coupled with each other and follow an inherent internal logic. 

Admittedly, 3DVLA introduces additional sequential modules. Therefore, we supplement a series of robustness experiments to evaluate the cumulative error, as shown in Table~\ref{tab:noise_robustness}. We conduct controlled inference-time perturbation experiments on two components: instance predictions and completion coordinates. Completion Noise is implemented by adding Gaussian noise to the predicted completion coordinates, where $a=1$ means the signal is entirely noise. Instance Noise is implemented by randomly replacing a proportion $a$ of the predicted instances with random instances sampled from other scenes. 

In fact, even without manually injected noise, the model's own intermediate predictions are not perfect: we manually evaluate 100 test scenes and observe that the instance prediction recall is approximately 95\%. The experimental results demonstrate that 3DVLA consistently achieves a higher success rate than the baseline under varying degrees of noise, proving that our proposed 3D modules are robust against imperfect intermediate geometric representations.

\subsection{Multi-seed Evaluation}
\label{sec:multi_seed}

To rigorously validate our performance deltas and rule out run-to-run variance, we conduct expanded evaluations using five independent random seeds for both the LIBERO-Plus and RoboTwin 2.0 benchmarks. 

As summarized in Table~\ref{tab:seed_libero} and Table~\ref{tab:seed_robotwin}, the results from repeated experiments with multiple random seeds continue to support our original conclusions. Moreover, we observe that 3DVLA can reduce variance to a certain extent. This suggests that injecting explicit 3D structural priors helps alleviate behavioral uncertainty under complex visual and geometric variations.

\subsection{Qualitative Video Comparisons}
\label{sec:video_comparisons}
We provide a qualitative demonstration of 3DVLA on diverse manipulation scenarios, including LIBERO-Plus, RoboTwin 2.0, and a real-world cloth folding task. The video shows that baseline VLA models often fail under challenging spatial conditions and severe occlusions, while 3DVLA achieves robust manipulation by leveraging 3D-aware representations and geometry reasoning.



\end{document}